\documentclass[conference]{IEEEtran}
\IEEEoverridecommandlockouts
\pdfoutput=1
\usepackage{cite}
\usepackage{amsmath,amssymb,amsfonts}
\usepackage{graphicx}

\usepackage{textcomp}

\usepackage{booktabs} 
\usepackage{color}
\usepackage{float}
\usepackage{mathtools,subfigure}
\DeclarePairedDelimiter{\abs}{\lvert}{\rvert} 
\usepackage{algorithm}
\usepackage{algorithmicx}
\usepackage{algpseudocode}
\usepackage{bm}
\usepackage{relsize}

\usepackage{url}  
\usepackage{graphicx}  
\usepackage{balance}

\def\BibTeX{{\rm B\kern-.05em{\sc i\kern-.025em b}\kern-.08em
    T\kern-.1667em\lower.7ex\hbox{E}\kern-.125emX}}
\begin{document}

\title{A Novel Multiple Classifier Generation and Combination Framework Based on Fuzzy Clustering and Individualized Ensemble Construction }

\author{\IEEEauthorblockN{Zhen Gao}
\IEEEauthorblockA{\textit{Department of Computer Science} \\
\textit{University of Texas at San Antonio}\\
San Antonio, United States \\
zhen.gao@utsa.edu }
\and
\IEEEauthorblockN{Maryam Zand}
\IEEEauthorblockA{\textit{Department of Computer Science} \\
\textit{University of Texas at San Antonio}\\
San Antonio, United States \\
maryam.zand@utsa.edu }
\and
\IEEEauthorblockN{Jianhua Ruan\textsuperscript{*} }
\IEEEauthorblockA{\textit{Department of Computer Science} \\
\textit{University of Texas at San Antonio}\\
San Antonio, United States \\
Jianhua.Ruan@utsa.edu }

}

\maketitle

\begin{abstract}
Multiple classifier system (MCS) has become a successful alternative for improving classification performance. However, studies have shown inconsistent results for different MCSs, and it is often difficult to predict which MCS algorithm works the best on a particular problem. We believe that the two crucial steps of MCS - base classifier generation and multiple classifier combination, need to be designed coordinately to produce robust results. 
In this work, we show that for different testing instances, better classifiers may be trained from different subdomains of training instances including, for example, neighboring instances of the testing instance, or even instances far away from the testing instance. 
To utilize this intuition, we propose Individualized Classifier Ensemble (ICE). ICE groups training data into overlapping clusters, builds a classifier for each cluster, and then associates each training instance to the top-performing models while taking into account model types and frequency. In testing, ICE finds the $k$ most similar training instances for a testing instance, then predicts class label of the testing instance by averaging the prediction from models associated with these training instances. 
Evaluation results on 49 benchmarks show that ICE has a stable improvement on a significant proportion of datasets over existing MCS methods. ICE provides a novel choice of utilizing internal patterns among instances to improve classification, and can be easily combined with various classification models and applied to many application domains.
\end{abstract}


\section{Introduction}

Multiple classifier system (MCS), including ensemble classifiers and mixture of experts, has established itself as an effective and practical solution to address challenges in supervised learning, such as functional complexity, insufficient training data, high dimensionality of feature space, and noise in training data, among others. Many excellent comprehensive reviews on MCS algorithms are available~\cite{rokach2010ensemble,kuncheva2004combining,oza2008classifier}. 

Learning a MCS usually includes two critical steps: base classifier generation, and multiple classifier combination, although sometimes the two steps are intrinsically integrated. Different MCS methods can be distinguished by how these two steps are performed. According to model generation strategies, existing MCS methods usually fall into one of the following two categories: random methods and deliberate methods. The former generates models by injecting random perturbations into the training data or training process~\cite{breiman1996bagging,breiman2001random}. In contrast, the latter attempts to generate multiple classifiers in a more systematic, principled way, e.g., by iteratively re-weighting the training instances with emphasis on previously misclassified instances, a technique known as boosting~\cite{freund1995desicion}, or by first clustering the training instances and then learning submodels from each cluster~\cite{vilalta2003class,kuncheva2000clustering}. According to model combination strategies,  MCS methods can also be grouped into two categories: voting-based and learning-based. Most popular ensemble methods (e.g., bagging and boosting) take a (weighted) voting from all models in the pool. Other methods attempt to learn a high-level model in order to determine which model(s) should be selected for the prediction task, or to learn a more complex function to combine the outputs of all models in the pool. Learning-based model combination algorithms include stacking, dynamic model selection, among many others~\cite{Stacking92,giacinto2001dynamic}.

Overall, ensemble approaches combining randomized model generation and voting (e.g. bagging and random forest) have been more successful / popular, probably due to their simplicity and less over-fitting. 
On the other hand, it has been shown that careful integration of deliberate models and learning-based model combination can be very effective on specific problem domains~\cite{Gashler2008}. In particular, empirical studies suggest that many classification problems consist of subdomains, which can potentially benefit from constructing and selecting submodels~\cite{jahid2014personalized,vilalta2003class,kuncheva2000clustering}. The challenge, however, lies in whether these subdomains can be corrected identified at training, and whether the  submodels can be correctly selected for individual cases at prediction time.

Here, we design a general MCS framework, Individualized Classifier Ensemble (ICE), with two key ideas. First, it constructs a large pool of submodels that have low bias when applied to appropriate instances. This is achieved by applying a strong learner (in contrast to the high-bias, low-variance models commonly used in a few ensemble methods) to individual overlapping clusters of instances that represent possible subproblems. Second, a simple yet effective, learning-free method is used to obtain different combinations of submodels for different testing instances. The learning-free nature of the method reduces the chance of selecting wrong models, therefore ensures that the combination of the selected submodels is better than, or at least no worse than, an average of all submodels. 

Experimental results on 49 datasets from different domains show that ICE consistently outperforms the competing methods. Furthermore, detailed component analysis shows that both steps of our algorithm have positive contributions as expected. In addition, analysis of the submodels can shed light on the internal structure of the problem, which can potentially be used to further increase  prediction performance, or to improve mechanistic understanding of the problem. The framework can be easily combined with existing classifiers and applied to many domains.



\begin{figure}[t] 
\centering
\includegraphics[width=.48\textwidth]{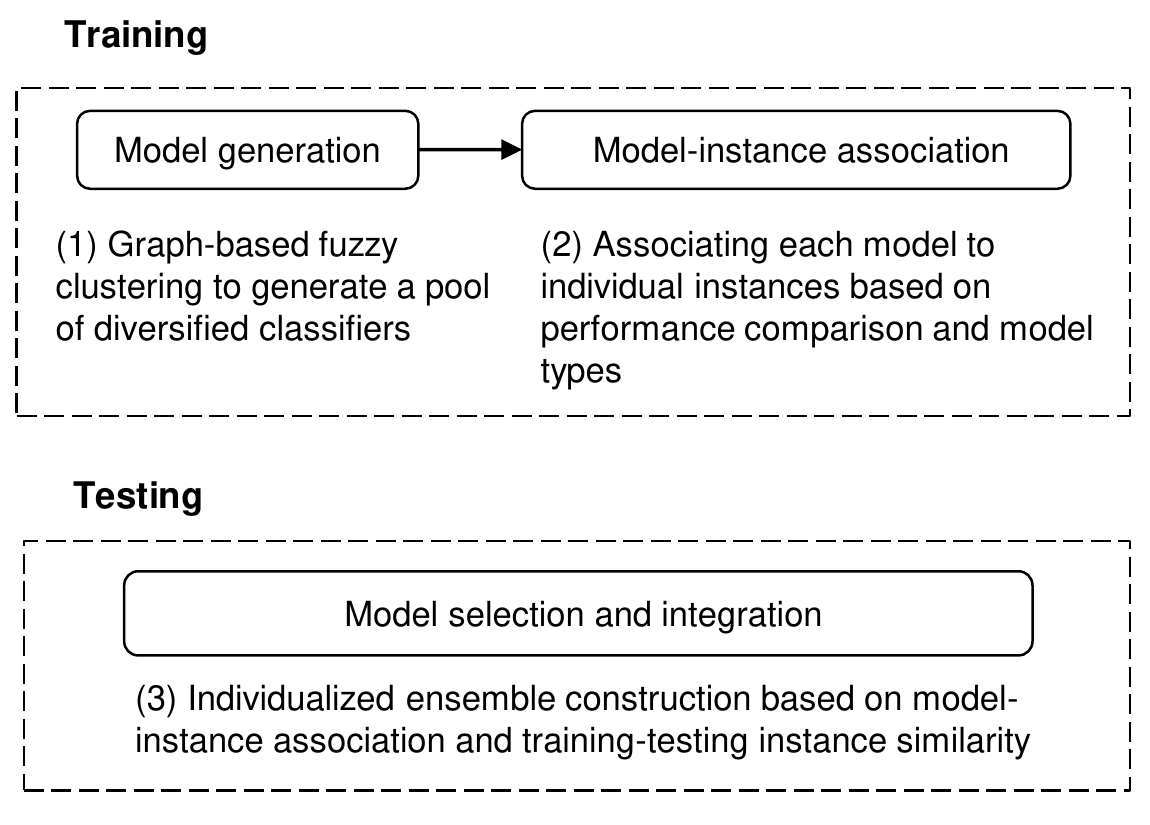}
\caption{\textbf{Overview of the ICE framework.}  }
\label{fig_framework}
\end{figure}

\section{Methods}


Fig.~\ref{fig_framework} shows a brief overview of ICE, which starts with generating a pool of diverse and subdomain-representative classifiers from subsets of training instances (Algorithm 1), obtained by a graph-based clustering method that can detect overlapping clusters (Algorithm 2). Then, these classifiers are associated with individual training instances based on their relative prediction performance on the instance, taking into account model types and frequency (Algorithm 3). In testing/prediction stage, the nearest neighbors of a test instance are identified from the training dataset and the classifiers associated with these neighboring instances are selected to form an ensemble for prediction (Algorithm 4). While the general ICE framework is flexible and the individual components can be re-designed with domain-specific information, several design principles are crucial and are discussed below. 

Source code and data are available at \url{https://github.com/ds-utilities/ICE}.


\begin{algorithm}
{    \small
    \caption{Training}
    \label{f_train}
    \begin{algorithmic}[1] 
        \Procedure{train }{ $\bm{X}$, $\bm{Y}$, $L$, $w$, $s$ } 
            \State $\bm{C} \gets $ CLUSTERING $ (\bm{X}, L)$
            \State $\bm{O} = \{o_i \}_{i=1}^L \gets \emptyset$
            \For{$c_j$ in $\bm{C}$} 
                \State $ o_j \in \bm{O}  \gets$ Train model on $c_j$ using base-classifier
            \EndFor
            \State $\bm{D} \gets  $  \small{ASSOCIATE} $ (\bm{X}, \bm{Y}, \bm{C}, \bm{O}, w, s )$
            \State \textbf{return} $\bm{D}$, $\bm{O}$
        \EndProcedure
    \end{algorithmic}
    }
\end{algorithm}

\begin{algorithm}
{\small
    \caption{Graph-based fuzzy clustering}
    \label{f_clustering}
    \begin{algorithmic}[1] 
        \Procedure{rwfclustering }{$\bm{X}$, $L$}
            \State $\bm{S}_{Q \times Q} \gets $ Euclidean dist. based inst. similarities on $ \bm{X}$
            \State $\bm{G}_{Q \times Q} \gets $ Each node keeps top $\lceil log_{10}Q \rceil$ Nbr. on $\bm{S}$
            \State $\bm{W}_{Q \times Q} \gets $ Random walk with restart on $\bm{G}$
            \State $\bm{T} = \left\langle t_j \right\rangle_{(L-1) \times 1} \gets \left\langle 0 \right\rangle_{(L-1) \times 1}$
            \State $t_1 \gets $ find the most connected node in $ \bm{W} $
            \For{$j$ in $2 ... (\bm{L}-1)$} 
                \State $t_j \gets $ find the common farthest node of $\bm{T}$
            \EndFor
            \State $\bm{R} = \left\langle r_{ij} \right\rangle_{Q \times Q} \gets $ keep top $ z $ edges in $\bm{W} $ 
            \State $\bm{C} = \{c_i \}_{i=1}^L \gets \emptyset$
            \For{$j$ in $1 ... (\bm{L}-1)$} 
                \State $c_j \gets$ find indices of $ r_{t_{j} \bullet } == 1$ 
            \EndFor
            \State $ c_L \gets 1 ... Q$ 

            \State \textbf{return} $\bm{C}$
        \EndProcedure
    \end{algorithmic}
}
\end{algorithm}

\subsection{Training}


\subsubsection{Basic notations}

We define a dataset of $Q$ training instances as $\bm{A} = \{(x_i, y_i) \}_{i=1}^Q$, where $x_i\in \bm{X}$ is an $R$ dimensional  feature vector and $y_i  \in \bm{Y}$ is the binary label of instance $i$. 
The clustering result on $\bm{X}$ is denoted as $\bm{C} = \{c_i \}_{i=1}^L$; $c_i$ is the $i$th cluster; $L$ is the total number of clusters. Here we designate the last cluster $c_L$ of $\bm{C}$ to be the whole set of instances. Without loss of generality, we assume the class labels are binary.

\subsubsection{Graph-based Fuzzy Clustering} 

As clustering can be subjective and unstable, we recommend generating a large number of relatively independent but overlapping clusters. In addition, each cluster needs to have a sufficient number of instances to learn a strong submodel for that subdomain. In our design, we use a graph-based clustering algorithm that chooses a set of furthest points to initiate a random walk process and use probability cutoffs to control cluster size (Algorithm~\ref{f_clustering}). 

The algorithm works as follows. We first calculate an instance-instance distance matrix on $\bm{X}$ by Euclidean distance and store it in $\bm{S}$. Then, we construct a \textit{KNN} graph $\bm{G}$ by keeping the top $\lceil log_{10}Q \rceil$ neighbors  for each node in $\bm{S}$. 
Afterwards, a random walk with a restart probability $p$ (default to 0.3 in this work) is performed on the \textit{KNN} graph $\bm{G}$ to obtain an affinity matrix, $\bm{W}$~\cite{RWR2006}. Next, a set of points, $\bm{T} = \left\langle t_j \right\rangle_{(L-1) \times 1}$, is identified as cluster centers: from $\bm{W}$, the  node with the largest total incoming probability,  $t_1$, is chosen as the center point of the first cluster; cluster centers for the other clusters are selected by finding the furthest node from the current center points. Finally, a probability cutoff is applied on $\bm{W}$ to identify direct neighbors of each cluster center as members of the cluster, such that the average cluster size is $z$ ($z=Q/3$ as default). We designate the last cluster $c_L$ of $\bm{C}$ to be the whole set of instances. A classifier is built using instances from each cluster. 


\subsubsection{Associating models to instances}

\begin{algorithm}
{ \small
    \caption{Associate instances with models by calculating the decision table}
    \label{f_dec_mat}
    \begin{algorithmic}[1] 
        \Procedure{associate }{ $\bm{X}$, $\bm{Y}$, $\bm{C}$, $\bm{O}$, $w$, $s$ } 
            \State $ \bm{P} = \left\langle p_{ij} \right\rangle_{Q \times L} \gets \left\langle 0 \right\rangle_{Q \times L} $
            \State $ \bm{E} = \left\langle e_{ij} \right\rangle_{Q \times L} \gets \left\langle 0 \right\rangle_{Q \times L} $
            \For{$c_j$ in $\bm{C}$}
                \State $ \left\langle p_{\bullet j} \right\rangle_{Q \times 1}  \gets$  Pred. $\bm{Y}$ with Cross Val. $(o_j, c_j, \bm{X})$
                \State $ \left\langle e_{\bullet j} \right\rangle_{Q \times 1} \gets \abs{\left\langle p_{\bullet j} \right\rangle_{Q \times 1} - \bm{Y}}$ 
            \EndFor

            \For{$\left\langle e_{i \bullet }\right\rangle_{1 \times L}$ in $\bm{E}$} 
                \State $e_{iL}  \gets$ $(e_{iL} - w) $
                \State $e_{ij}   \gets$ $(e_{ij} - s)$ if $x_i \in c_j$
            \EndFor

            \State $ \bm{D} = \left\langle d_{ij} \right\rangle_{Q \times L} \gets \left\langle 0 \right\rangle_{Q \times L} $
            \For{$d_{ij}$ in $\bm{D}$} 
                \State $d_{ij}   \gets$ 1 if $e_{ij}\leq e_{iL}$, 0 otherwise
            \EndFor
            \State \textbf{return} $\bm{D}$

        \EndProcedure
    \end{algorithmic}
}
\end{algorithm}

Incorrect model selection can significantly degrade the performance of the algorithm compared to simply averaging all submodels. When the number of training instances is relatively small, supervised learning based model selection tends to overfit. Therefore, we propose a robust learning-free method (Algorithm 3), which performs model-instance association at training time and KNN-based model selection at prediction time. Importantly, the model-instance association step takes a Bayesian approach by using different cutoffs for different types of submodels, which reflects their frequency in the pool and the probability for them to outperform other types of submodels. 

Formally, given the clustering result on instances, $\bm{C} = \{c_i \}_{i=1}^L$, where $c_L$ is the whole set of instances, the corresponding set of models built on the clusters by a base learner (e.g., SVM) is denoted as $\bm{O} = \{o_i \}_{i=1}^L$. Here we call a model $o_i, i \in [1,  L-1]$ as a `$partial$' model, since each model is built on a subset of the training instances, and, we call model $o_L$ as the `$whole$' model, which is built on the whole set of instances. The class probabilities predicted by all models are stored in $ \bm{P} = \left\langle p_{ij} \right\rangle_{Q \times L} $; $p_{ij}$ is the predicted class probability for instance $i$ by model $j$; $p_{iL}$ is the prediction probability for instance $i$ by model built on the whole set of training instances. Note that  if instance $i$ is NOT a member of cluster $c_j$ (in which case, we call model $o_j$ to be a `$remote$' model of instance $i$), the model is directly used to predict $p_{ij}$ for instance $i$; on the other hand, if instance $i$ is a member of cluster $c_j$ (in which case we call model $o_j$ a `$local$' model of instance $i$), the value $p_{ij}$ is obtained by 10-fold cross-validation using instances in this cluster. This process ensures that the performance evaluation used for model-instance association is not inflated, as an instance is never evaluated by a model that used the instance in training. Importantly, by not having any designated validation dataset, we are able to keep as many instances as possible for training, an important feature for small training data.

The prediction error table, $ \bm{E} = \left\langle e_{ij} \right\rangle_{Q \times L} $ is derived from $\bm{P}$; $e_{ij} = \abs{p_{ij} - y_i}$ is the prediction error for instance $i$ by model $o_j$. Each row of $\bm{E}$, $e_{i \bullet }$, represents the prediction error of different models on instance $i$. 
Given the empirical results that $local$  models usually work slightly better than $whole$  model and $remote$ models, as well as the fact that there are more $remote$  models than $local$  models in the pool, we introduce two parameters to easily balance the proportion of $local$, $whole$ and $remote$ models in the ensemble: $w$ as the advantage score of the $whole$ model, and $s$ the advantage score of each $local$ model. Usually $s > w > 0$ to promote the inclusion of $local$  models and demote $remote$  models, unless the error in a remote model is significantly smaller than in the $whole$  model. Each row of $\bm{E}$ is adjusted such that $e_{iL} \gets (e_{iL} - w) $, and, $e_{ij} \gets (e_{ij} - s) $ if $x_i \in c_j$. Then, the decision table, $ \bm{D} = \left\langle d_{ij} \right\rangle_{Q \times L} $, $d_{ij} \in \{1, 0\}$, where $d_{ij} = 1$ indicates association between model $o_j$ and instance $i$, is derived from the error table $\bm{E}$, 
 by 
 \[
    d_{ij} = 
\begin{cases}
    1, & \text{if }  e_{ij}\leq e_{iL} \\
    0, & \text{otherwise}
\end{cases}
\]

\subsection{Testing / prediction}

\begin{algorithm}
{\small
    \caption{Testing}
    \label{f_predict}
    \begin{algorithmic}[1] 
        \Procedure{Predict }{ $x_t$, $\bm{X}$, $\bm{D}$, $\bm{O}$, $N$, $\alpha$, $\beta$ } 
            \State $K^{nb}= \left\langle k_i \right\rangle_{N \times 1}  \gets $ select $N$ nearest Nbr. of $x_t$  
            \State $\bm{O}^{nb} \gets \emptyset$
            \For{$k_i$ in $K^{nb}$} 
                \State $J \gets \left\langle d_{k_{i}\bullet } \right\rangle_{1 \times L} = 1 $ 
                \State $\bm{O}^{nb} \gets \bm{O}^{nb} \cup \bm{O}_{J} $
            \EndFor
            \State $p^{whole} \gets $ Predict by base-classifier $(o_L, x_t)$
            \State $M \gets length(\bm{O}^{nb}) $
            \State $\bm{P}^{partial} = \left\langle p^{partial}_i \right\rangle_{M \times 1} \gets \emptyset$
            \For{$o_i$ in $\bm{O}^{nb}$} 
                \State $p^{partial}_i$  $ \gets$  Predict by base-classifier $(o_i, x_t)$ 
            \EndFor
            \State $p^t$ $\gets$ Equation \ref{eq1}$(\bm{P}^{partial}, p^{whole}, M, N, \alpha, \beta )$ 
            
            \State \textbf{return} $p^t$
        \EndProcedure
    \end{algorithmic}
    }
\end{algorithm}

For a test instance $x_t$, ICE first finds its $N$ nearest neighbors from the training dataset, then predicts its class label $y_t$  by averaging the class probabilities predicted by the models associated with the neighbor training instances (Algorithm \ref{f_predict}). Formally, the \textit{PREDICT}() algorithm first selects $N$ nearest neighbors of $x_t$ from $\bm{X}$, and stores the indices of the neighbor instances in $K^{nb}= \left\langle k_i \right\rangle_{N \times 1}$. Then, for each neighbor instance $k_{i}$, the algorithm looks up in the corresponding decision table $d_{k_{i}\bullet }$ to find the models associated with the neighbor instance, and stores the associated `$partial$' models of $x_t$ in $\bm{O}^{nb}$. The number of `$partial$' models in $\bm{O}^{nb}$ is denoted as $M$. Note that although $o^{nb}_i \in \bm{O}$, $\bm{O}^{nb}$ is not a subset of $\bm{O}$, since $\bm{O}^{nb}$ may contain duplicated models. Then we denote $\bm{P}^{partial} = \left\langle p^{partial}_i \right\rangle_{M \times 1}$ as the `$partial$' model predictions, and each $p^{partial}_i$ is predicted by $o_i$ on $x_t$. The predicted class probability by the whole model is denoted as $p^{whole}$. Then the predicted class probability of $x_t$ is calculated by:

\begin{equation} \label{eq1}
p^t=\frac{\mathlarger{\sum}_{i=1}^{M}{p_i^{partial}+(\alpha M + \beta N)\cdot p^{whole} } } {(\alpha + 1) M + \beta N},
\end{equation}

where $\alpha$ is the parameter to balance the weight of `$partial$' models and the `$whole$' model; $\beta$ is the parameter to adjust the weight of `$whole$' models based on the number of top neighbors to ensure at least one $p^{whole}$ will be used in case there is no `$partial$' model. 

In our experiments, $\alpha$ and $\beta$ are both set to 1 and N is set to 5, except in cases that we vary them to analyze the contribution of different  components and the robustness of our algorithm's performance.

\subsection{Relationship with Existing MCS Methods}

ICE differs from most existing ensemble methods significantly in both model generation and model combination. Popular ensemble methods such as Bagging and Random Forest generate submodels using random subsets of data, and combine them using voting. In order for these methods to work effectively, a large number of submodels is needed to reduce overall bias. In contrast, ICE generates submodels to deliberately increase model diversity by clustering training instances. 
A carefully designed model-instance association algorithm helps identify the best ensemble for individual instances at prediction time. On the other hand, boosting generates submodels that focus on different groups of training instances, where grouping of instances is done implicitly by iterative re-weighting and therefore lack a global view of instance space. In addition, since there is no model selection at prediction time, boosting tend to overfit in the presence of noisy training instances.

Mixtures of experts is a class of neural network models attempting to simultaneously learn multiple submodels as well as a gating function that assigns each instance to one or more submodels~\cite{jacobs1991adaptive,jordan1994hierarchical} .  With similar idea, several methods use clustering as a preprocessing step for classification~\cite{vilalta2003class,kuncheva2000clustering}. 
These algorithms force each instance to be in a disjoint cluster, which reduces the number of instances at training time. In addition, prediction is done only by cluster-specific models so the cost of incorrect model selection is high. Empirical results presented in the original papers show mixed performance when compared to other MCS algorithms~\cite{jacobs1991adaptive,vilalta2003class,kuncheva2000clustering}.  

Finally, a series of methods have been developed recently under the common name `dynamic model selection' \cite{cruz2015meta,ko2008dynamic,ko2008dynamic,kurzynski2016multiclassifier,woloszynski2012measure,woods1997combination,giacinto2001dynamic,giacinto1999methods}. These approaches take an ensemble of base classifiers (e.g, from bagging), then attempt to learn a high-level classification model using, for example, instance-instance similarities and model-model correlations, as input features. While conceptually appealing, these methods tend to overfit and have poor performance when training data is limited. In our opinion, the marriage between random model generation and learning-based model combination is a poor choice,  since the relatively small number of random models (compared to the possible number of instance combinations) does not guarantee that there is necessarily any \textit{predictably} better submodel than a simple average of all submodels. 


\section{Results and Discussion}

\subsection{Data and Experimental Setup}

\begin{figure}[t]
\centering
\includegraphics[width=0.48\textwidth]{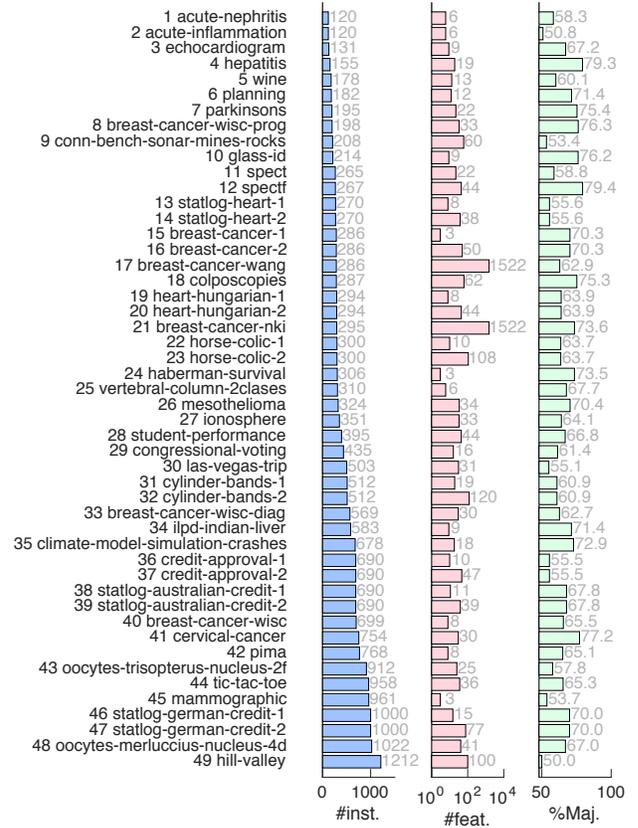}
\caption{\textbf{Characteristics of datasets used in evaluation. } The three columns show number of instances, number of features and percentage of majority class, respectively. }
\label{fig_simple_stat}
\end{figure}

Characteristics of the 49 benchmark datasets are shown in Figure \ref{fig_simple_stat}. The datasets are collected from UCI machine learning repository and Kaggle Dataset for binary classification, with number of instance between 100 and 3,000, number of features between 3 and 1500, and percentage of majority class ranging from 50\% to 77\%. A total of 42 datasets from UCI and 23 datasets from Kaggle meet the criteria (18 of which appeared in both repositories). 
In addition, we add two cancer-related datasets - breast-cancer-nki ~\cite{van2002gene} and breast-cancer-wang ~\cite{wang2005gene}. 
The data preprocessing mainly follows~\cite{fernandez2014we}, which includes a $Z$-Score transformation based normalization. For nine datasets with nominal features 
we use two different methods to handle nominal features: (i) removing nominal features (denoted with suffix `-1' in Figure 2), (ii) using One-Hot encoding (denoted with suffix `-2' in Figure 2). Since all features in dataset `tic-tac-toe' are nominal, this dataset only has the One-Hot encoding version. Data and source code are available at \url{https://github.com/ds-utilities/ICE}.

Performance of each classification method is evaluated by 10-fold cross validation and measured by AUC. To facilitate a simple and fair evaluation, we use common parameter values for ICE on all datasets. The number of overlapping clusters,  $L$, is set to 100, which while not ideal for all data sets, makes evaluation easier. The advantage scores for `$whole$' model and `$local$' model are set to $w = 0.4$ and $s = 0.5$ respectively; this reflects the empirical observation that $local$ models usually have better performance than the other two types of models, and there are many $remote$ models so a higher cutoff score is needed for a $remote$ model to be associated with an instance. In prediction stage, the number of top neighbors parameter $N$ is set to 5; the parameter $\alpha$ and $\beta$ are both set to 1 for an overall balanced `$partial$' and `$whole$' models in the final weighting of prediction. The base model in the evaluation is linear-SVM with the regularization parameter $C$=1 for ICE and comparison methods Bagging and AdaBoost. Bagging and AdaBoost use 100 bags and 100 iterations respectively.
It is worth noting that these parameters are chosen intuitively without extensive tuning. Parameter analysis results show that the performance of ICE is robust with regarding to a wide range. 


\begin{figure}[!tbp]
\centering
\includegraphics[width=3.3in]{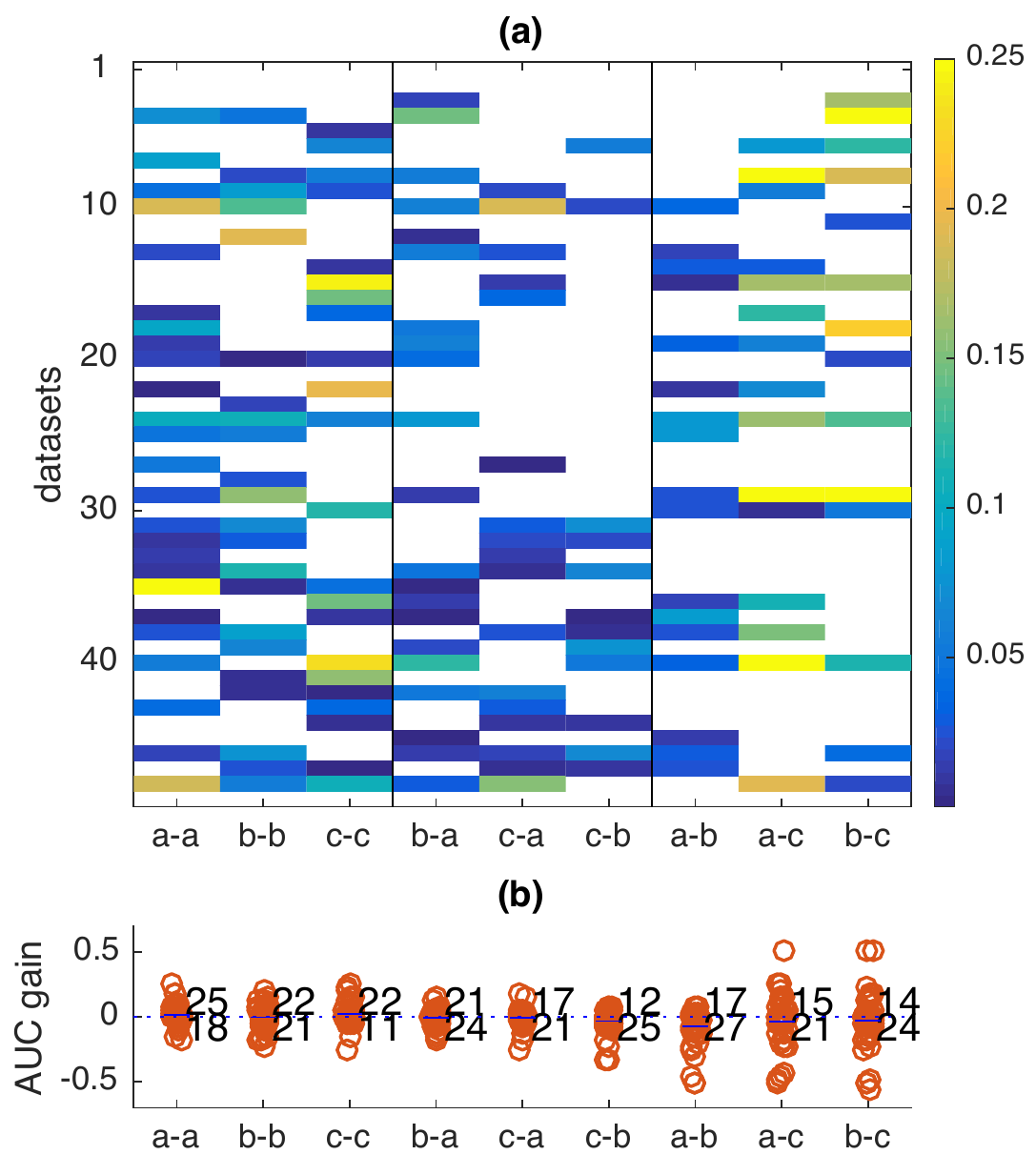}
\caption{\textbf{AUC gain of $partial$ models over $whole$ model.} (a) Color scale represents AUC gain; AUC gain $\le 0$ is removed to emphasize on potential benefits of using subsets of training instances. (b) each node represents the AUC gain on a dataset using a cluster of training data to predict a testing cluster. The two numbers on the right of each column are the number of datasets with AUC gain $>0$ and AUC gain $<0$.  }
\label{fig_vali}    
\end{figure}

\subsection{Empirical Evidence Supporting Cluster-Based Ensemble Classification}

To verify our assumption that, for each testing instance, some subset of training instances may provide a better classification model than the whole set of training instances, we perform a simple experiment as follows: first, each dataset is clustered into three disjoint clusters using \textit{k}-means. We denote the clusters as cluster-\textit{a}, \textit{b} and \textit{c} respectively, with their cluster size decreasing. Then using instances in each cluster for cross-testing: we compared the prediction AUC for each cluster using instances from cluster \textit{a}, \textit{b}, \textit{c} or the whole dataset, respectively, as training data. We adopted notation \textit{a}-\textit{b} to denote the situation where we use the cluster \textit{a} trained model to make predictions on cluster \textit{b} instances.

To have a fair evaluation, when using a larger cluster to predict a smaller cluster, we randomly select the same number of instances from the larger cluster as the size of the smaller cluster to be the training data; when use a smaller cluster to predict a larger cluster, we use all instances in the smaller cluster and randomly select some instances from the larger cluster (making sure that they are not in the fold of testing) to be the training instances, such that the total number of training instances is the same as the number of instances in the larger cluster.

From Figure \ref{fig_vali}, with only three disjoint clusters, in more than 80\% of the datasets, at least one of the $local$  models can outperform the $whole$  model (Figure \ref{fig_vali}a and b, columns \textit{a}-\textit{a}, \textit{b}-\textit{b} and \textit{c}-\textit{c}). Interestingly, while in general the $remote$  models do not perform well, some of them have the largest performance gain compared to the $whole$  model (column \textit{a}-\textit{c} and \textit{b}-\textit{c}). Collectively, this experiment shows the potential benefit of using a cluster of instances to improve prediction accuracy. On the other hand, the results also signifies the importance to predict, for each test instance, whether $partial$  models (and which) should be used.


\begin{table*}[!htbp]%
\centering
\caption{\textbf{AUC of ICE and competing methods on 49 datasets}   }
\includegraphics[width=1.2\textwidth, angle=90]{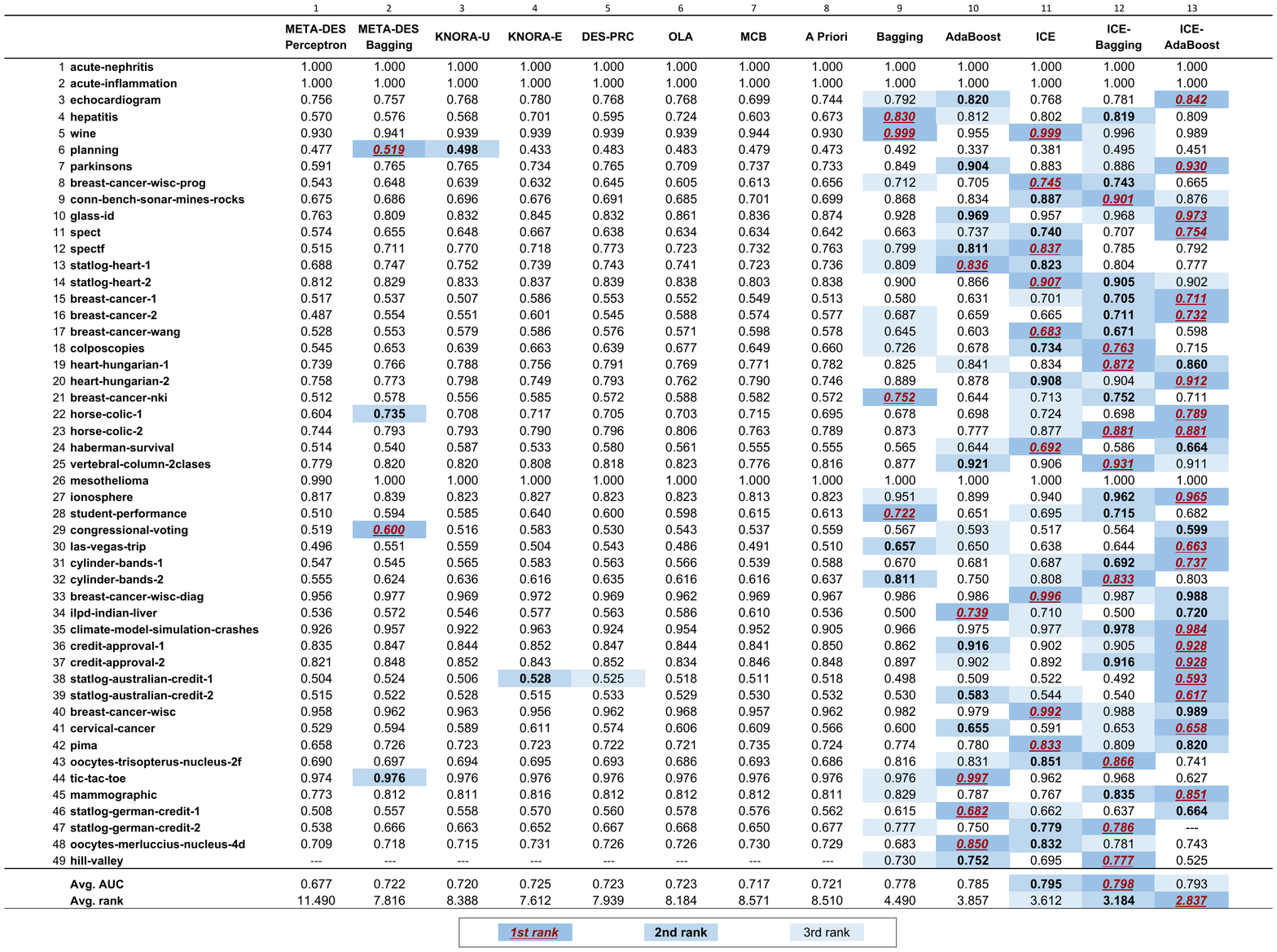}
\label{fig_ICEvsMetaDesMain}
\end{table*}

 \begin{figure}[t]
 \begin{center}
 \includegraphics[width=3.5in]{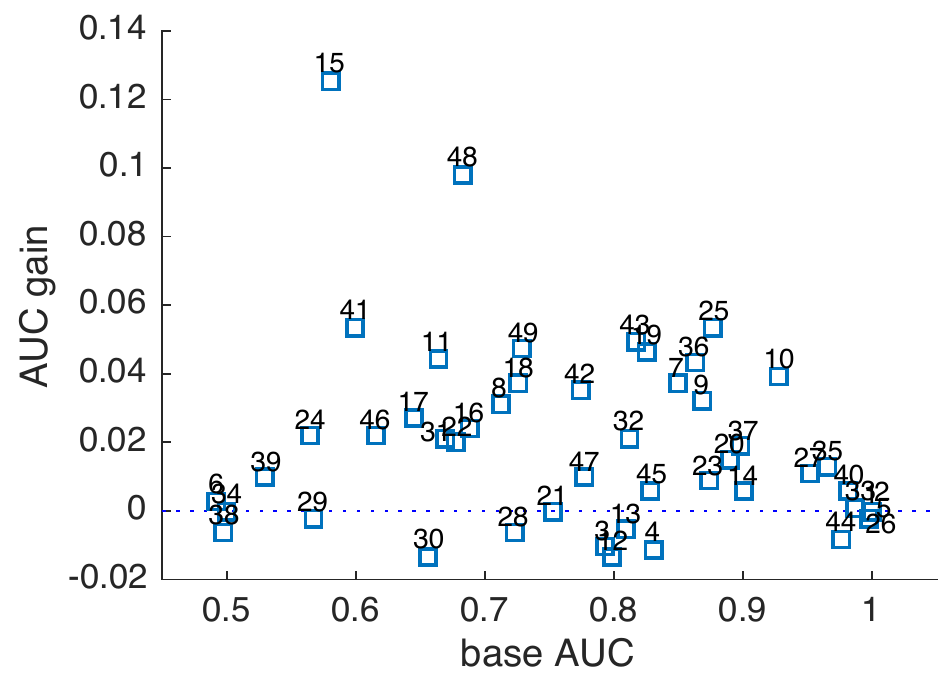}
 \caption{\textbf{ICE outperforms the corresponding benchmark classifier on most datasets.} The ID-name mapping of datasets is shown in Figure 1. $L=100$; $s = 0.5$; $w = 0.4$; $N = 5$;  $\alpha = 1$; $\beta = 1$. ICE wins on 37 out of 49 datasets (\%75.5).}
 \label{fig_main}
 \end{center}
 \end{figure}


\subsection{ICE Outperforms Existing MCS Algorithms}


Figure \ref{fig_main} shows that ICE outperforms the corresponding Bagging classifier on most datasets, and, suffers from only minor performance loss on a few datasets. Notably, ICE uses less than 100 base models - on average 45 models per prediction. ICE may still have room for improvement on failed datasets by parameter tuning and improved clustering methods. Understandably, ICE tends to have less performance gain on datasets with fewer instances, such as on datasets 1 to 6, since ICE needs more enriched instance information for a meaningful clustering. From another perspective, ICE will have advantage on datasets with more instances and with more complex instance structure.

Table 1 shows the complete AUCs of three versions of ICE (with SVM, Bagging and AdaBoost as the base model) on 49 benchmark datasets compared to multiple MCS methods, including Bagging, Adaboost, and seven dynamic model selection approaches. META-DES ~\cite{cruz2015meta} has two versions in this evaluation, using Perceptron (the base classifier choice of the original META-DES paper) and Bagging (comparable with Bagging and ICE-Bagging) respectively. The base classifier is Bagging for the other six dynamic model selection methods - KNORA-U ~\cite{ko2008dynamic}, KNORA-E ~\cite{ko2008dynamic}, DES-PRC ~\cite{kurzynski2016multiclassifier,woloszynski2012measure}, OLA ~\cite{woods1997combination}, MCB ~\cite{giacinto2001dynamic} and A Priori ~\cite{giacinto1999methods}, which is the suggested setting plus SVM to make comparable with other methods. We use the suggested parameters for dynamic model selection approaches ~\cite{cruz2018dynamic}. 

As shown, all three versions of ICE have better performance than the other methods. The performance gain of ICE over Bagging can be attributed to the use of specifically generated models for subproblems and individualized model association and selection step. Comparing AdaBoost to ICE, both models attempt to produce subdomain-specific classifiers; however, AdaBoost always uses the same ensemble of all submodels for all instances, which reduces the potential performance gain provided by the submodel-specific models. Therefore, ICE-Adaboost and even ICE-SVM perform better than AdaBoost in general. More over, ICE outperforms the seven dynamic selection methods. Each of the dynamic selection  methods has unique contributions on model selection or integration. However, none of them focuses on deliberately generating models for specific subproblems as the fuzzy clustering that ICE uses. In addition, the unique instance-model association of ICE can utilize all training instances, comparing to dynamic selection methods such as META-DES, which separates training data into META learning and dynamic selection datasets, therefore lead to more data loss and weaker base classifiers.  As discussed earlier in Section 2E, learning the best combination of multiple randomly generated models can be a daunting task when the amount of training data is limited. 

\subsection{Randomized Control Analysis Reveals The Effectiveness of Different Components of ICE}

\begin{figure}[t]
\centering
\includegraphics[width=3.4in]{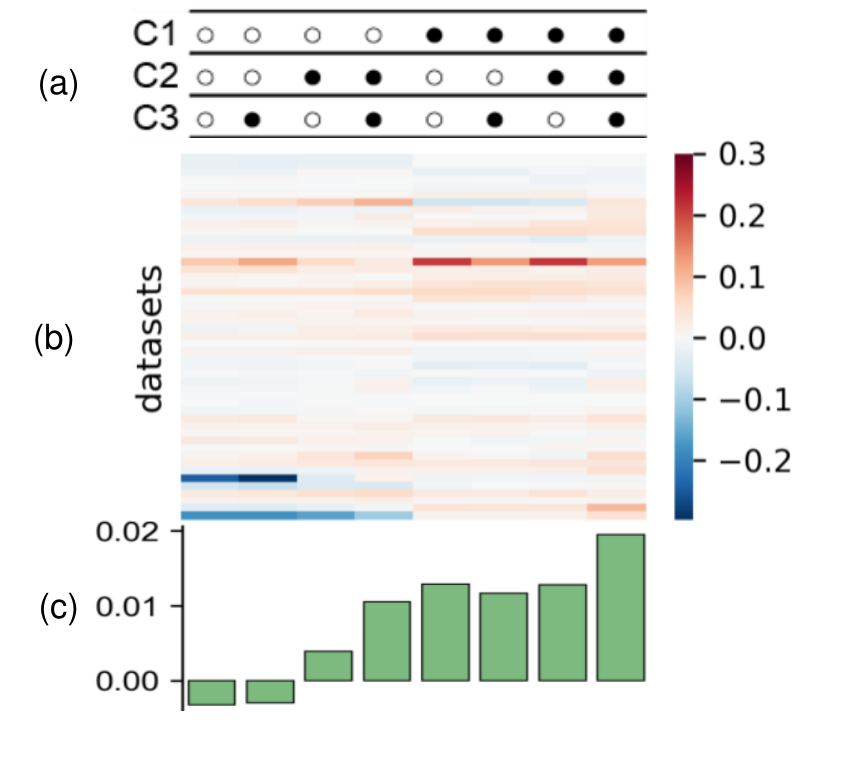}
\caption{\textbf{Component analysis of ICE.} Each column indicates a randomized control experiment.  \textbf{(a)} Marker `$\bullet$' and `$\circ$' represent standard component and random control.  \textbf{(b)} Color indicates the AUC gain of ICE over Bagging.  \textbf{(c)} Each bar shows the average AUC gain of ICE over Bagging.}
\label{fig_bitComp}
\end{figure}

To understand the impact of the three components of ICE (C1: fuzzy clustering based model generation, C2: instance-model association, and, C3: KNN-based model selection), we perform a randomized control experiment, where one or more of the components is replaced with comparable, randomized procedures. To randomize C1, the fuzzy clustering is replaced by bootstrapping instances , where the bags are made the same size as in the fuzzy clusters, therefore resulting in a slightly modified version of Bagging. To randomize C2, the decision table is shuffled row-wise, destroying the association of models to instances. Finally, to randomize C3, KNN is replaced with random selection of instances. Note that randomizing C2 or C3 (or both) are expected to have similar impact on the algorithm, which will essentially perform random model selection (and in most cases will choose many more models than real ICE due to independence of different rows of the randomized decision table). 

Figure~\ref{fig_bitComp} shows the performance of ICE  with different components randomized. Here, in order to show the effectiveness of each component of ICE, the parameter $\alpha$ and $\beta$ are set to 0, effectively eliminating `$whole$' model. Not surprisingly, when both the model generation and model selection components of ICE are randomized (columns 1-3 in Figure~\ref{fig_bitComp}a), its performance becomes similar to that of Bagging. On the other hand, when only one component is randomized (columns 4-7), ICE can still perform better than standard Bagging, although not as effective as the complete ICE algorithm (column 8), indicating that both components of ICE played a role in effective learning. 

Interestingly, with only C1 randomized, our algorithm is conceptually similar to dynamic model selection ~\cite{cruz2015meta}, except that we replaced their learning-based model selection with simple KNN-based model selection. The fact that this version of ICE still outperforms dynamic model selection suggests that, with limited training data, KNN-based model selection can have more robust performance than learning-based model selection.  In addition, when C2 or C3 (or both) are randomized but C1 is not randomized (column 5-7), our algorithm is conceptually similar to bagging, except that the models in the ensemble are based on clusters of instances instead of random selection of instances. As shown, this version of ICE has significant performance gain over Bagging, suggesting that, at least in these datasets, clustering-based model generation, which implicitly diversifies the models, can be better than randomized model generation.

\begin{figure}[t]
\begin{center}
\includegraphics[width=3.5in]{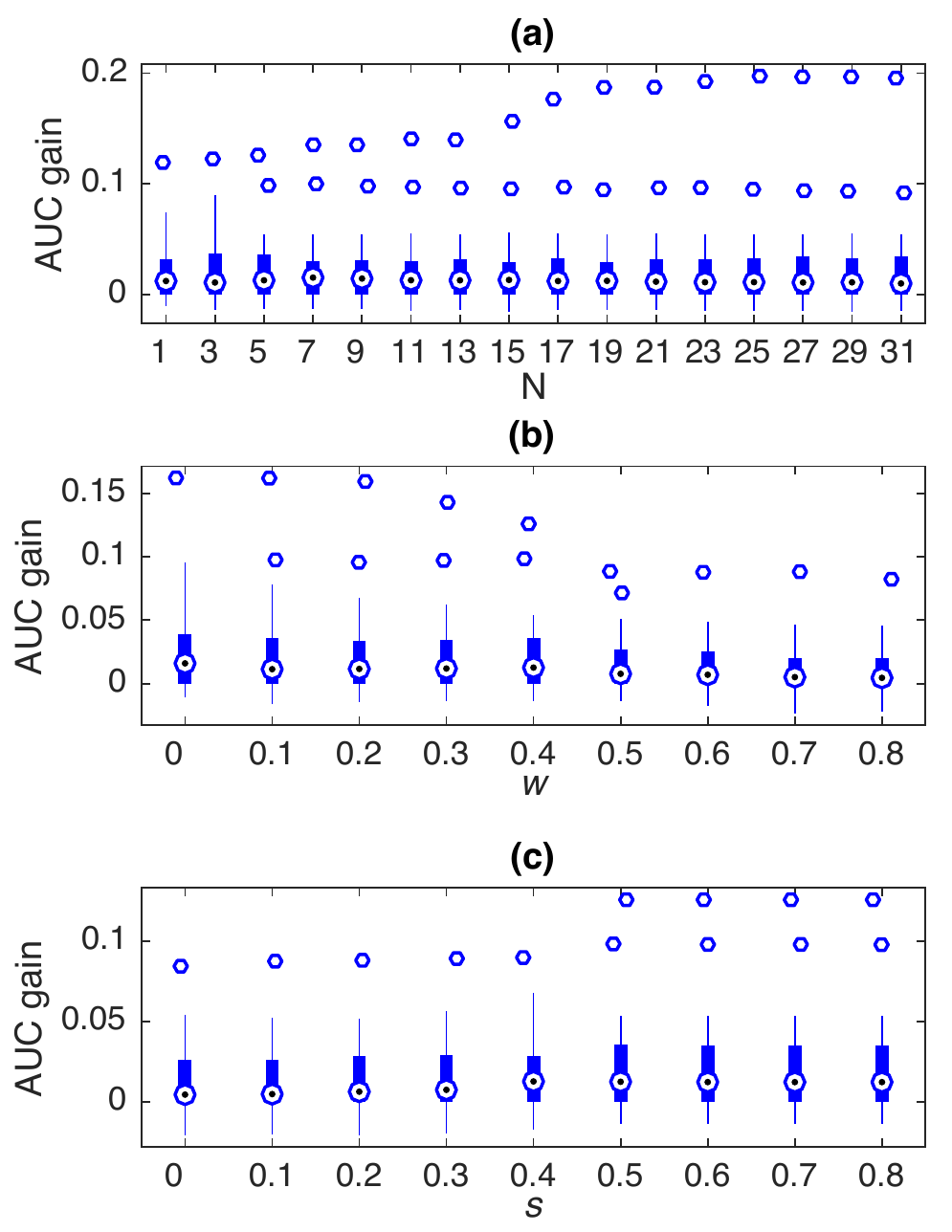}
\caption{\textbf{ICE performs in a stable manner across the wide range of parameter space.}  \textbf{(a)} AUC gain varies as a function of $N$, number of nearest neighbors for model selection. Here $w$ = 0.4, $s$ = 0.5. \textbf{(b)} AUC gain varies as a function of $w$. Here $N$ = 5, $s$ = 0.5. \textbf{(c)} AUC gain varies as a function of $s$.  Here $N$ = 5, $w$ = 0.4.}
\label{fig_para_grid}
\end{center}
\end{figure}

\subsection{Performance of ICE is Robust in a Wide Range of Parameter Space}
Figure \ref{fig_para_grid} shows the results of ICE using a wide range of parameters - $N$: number of neighbors per testing instances in prediction; $w$: the weight advantage of the base whole model in model-instance association; $s$: the weight advantage of the self-model in model-instance association. In this analysis, the parameter $\alpha$ and $\beta$ are both set to 1 to balance $whole$ and $partial$ models.

Figure \ref{fig_para_grid}a shows that the number of nearest neighbors used in model selection  has only slight impact on AUC gain on average across all 49 datasets. The recommended setting of $N$ is 5 to 10 for a balanced running speed and accuracy. ICE works best when there are strong patterns in the dataset. If ICE does not have a significant gain over Random Forest (RF) on a center dataset, a larger $N$ setting will make ICE more stable and closer to bagging. ICE still has a large room of improvement on specific dataset by using more suitable fuzzy clustering algorithm, which is one of our future work.

Figure \ref{fig_para_grid}b and Figure \ref{fig_para_grid}c shows the robust performance of ICE with respect to parameter $w$ and $s$. A general insight of $w$ and $s$ is to set s slightly larger than $w$, such as $s$ = 0.5, $w$=0.4. The parameter $\alpha$ and $\beta$ are quite simple to choose. Set both $\alpha$ and $\beta$ to 1 will lead to a decent result for most of cases; try to set both $\alpha$ and $\beta$ to 0 if there are strong clusters within the dataset, and the extreme localized classifiers may have an advantage over the basic to-go choice where $\alpha = \beta = 1$.

In addition, it is worth noting that the parameters used in the experimental setup have not been tuned for individual dataset in this study. There is a potential to perform model tuning on each dataset for even more improved performance.

\subsection{ICE Significantly Improves Random Forest Performance}
\begin{figure}[t]
\begin{center}
\includegraphics[width=3.5in]{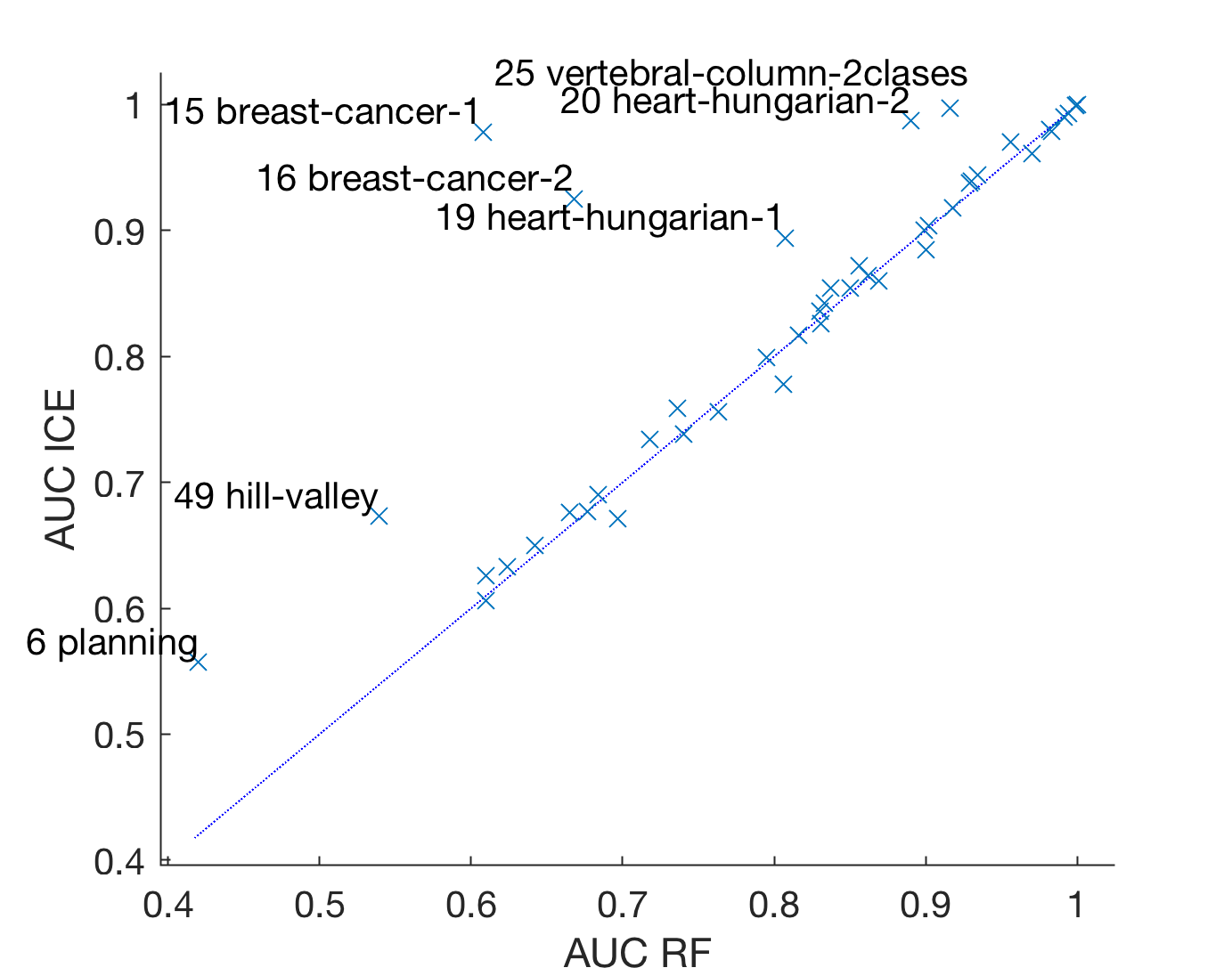}
\caption{\textbf{ICE significantly improves the performance of  random forest on seven datasets.} }
\label{fig_scatter_RF}
\end{center}
\end{figure}

We further perform an extreme comparison between ICE (using Random Forest with 100 trees as the base classifier) and Random Forest with 10,000 trees. Random Forest (RF) is well known for its stable high performance with almost tuning-free design, and is well positioned to be a benchmark classifier. As shown in Figure \ref{fig_scatter_RF}, ICE significantly improves the performance of  Random Forest; ICE wins or ties over RF on 36 out of 49 datasets (74\%), and has minor performance loss on 13 datasets. The $t$-test $p$-value of gain = 0.018, which is significant ($p$-value$<0.02$), and, there are 7 datasets (highlighted on Figure \ref{fig_scatter_RF}) with AUC gain over 8\% (among these, ICE has AUC gain over 13.4\% on 4 datasets), while no dataset with AUC loss over 3\%. Note that ICE only uses on average 47 models per prediction, much fewer comparing to 10,000 trees by the RF classifier. Moreover, RF easily reaches its performance limits as the number of trees grows, while ICE has a much larger room of improvement as the number of submodels increases. Performance of ICE can be further improved by increasing the number of fuzzy clusters (submodels) or using more suitable clustering methods.

The performance gain of ICE over RF can be attributed to the use of specifically generated models for subproblems and individualized model association and selection step. 
Interestingly, the AUC gain of ICE is correlated with the result from Figure \ref{fig_vali}a - the 7 highest-scoring datasets by ICE on Figure \ref{fig_scatter_RF} have on average 4.3 `$partial$' models winning the `$whole$' model, while this statistic is only 2.6 for the other datasets; the average AUC gain by  `$partial$' models of ICE on these 7 datasets is 0.077, while it is 0.057 for the other datasets. This results not only further validates our intuition of using `$partial$' models to improve classification performance, but also suggests that the performance of ICE can be partially predictable based on dataset characteristics, which is a very important feature in practice.

\subsection{Classification Improved by Accurately Predicting `Hard' Instances with `$partial$' Models}

\begin{figure}[t]
\centering
\includegraphics[width=3.5in]{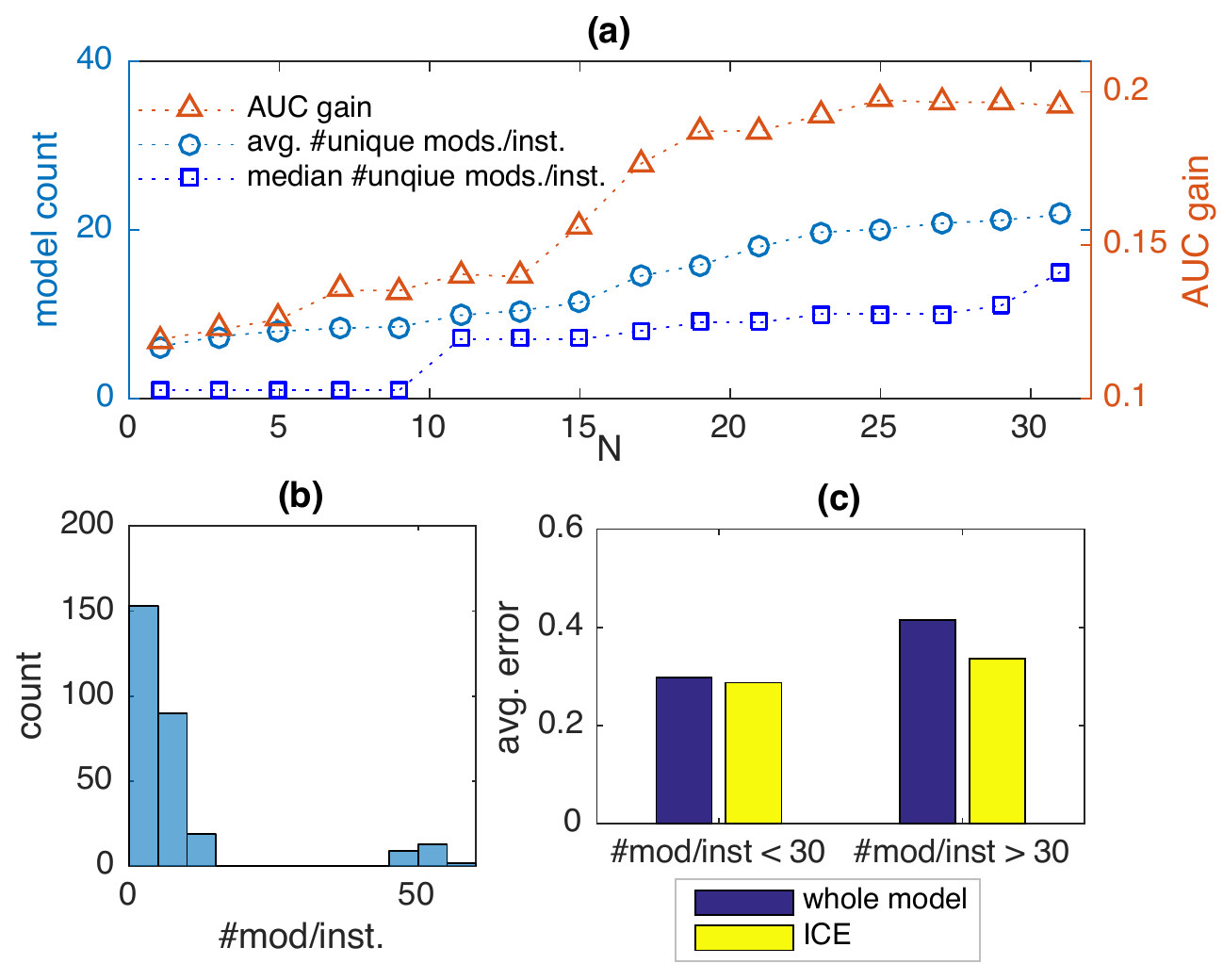}
\caption{\textbf{Analysis of ICE submodels on a breast cancer dataset.} \textbf{(a)} AUC gain of ICE varies as a function of model counts per instance. \textbf{(b)} distribution of number of unique `$partial$' models per instance. \textbf{(c)} Predictions by ICE have a lower error rate than Bagging on instances with $>$30 models.}
\label{fig_case_breast_cancer}
\end{figure}

ICE has a stable AUC gain on most of datasets over a large range of parameter variation, and the dataset with one of the most dramatic improvement using ICE is the 15-breast-cancer-1 dataset. As shown in Figure \ref{fig_case_breast_cancer}a, as the parameter $N$ increases, the average number of models per instance also increases and the performance of ICE continues to increase, reaching a plateau after $N \ge 25$. In addition, analysis of the models used by each test instance of ICE shows an interesting bimodal distribution: most of the test instances (262 out of 286 cases) use less than 20 models (mostly $local$  models); in contrast, a few instances (24 cases) use more than 40 unique models (including both $local$ and $remote$  models) (Figure \ref{fig_case_breast_cancer}b), which are presumably the more difficult instances that are hard to be clustered and/or classified.

Comparing  the performance difference on these two groups of instances, we can see that ICE has a much lower prediction error when compared to the`$whole$' model on instances with $>$30 models by ICE (Figure \ref{fig_case_breast_cancer}c). The $t$test $p$-value of the error differences between the`$whole$' model prediction and ICE prediction on instances with $>$30 models (24 cases) is significant ($8.98 \times 10^{-6}$). This result demonstrates that different instances should be treated differently on this dataset, and the ICE algorithm shows a potential way of separating and treating these different instances.

\subsection{AUC Gain of ICE has a Strong Correlation with the Data-Decision Table Similarity}

\begin{figure}[t]
\begin{center}
\includegraphics[width=3.5in]{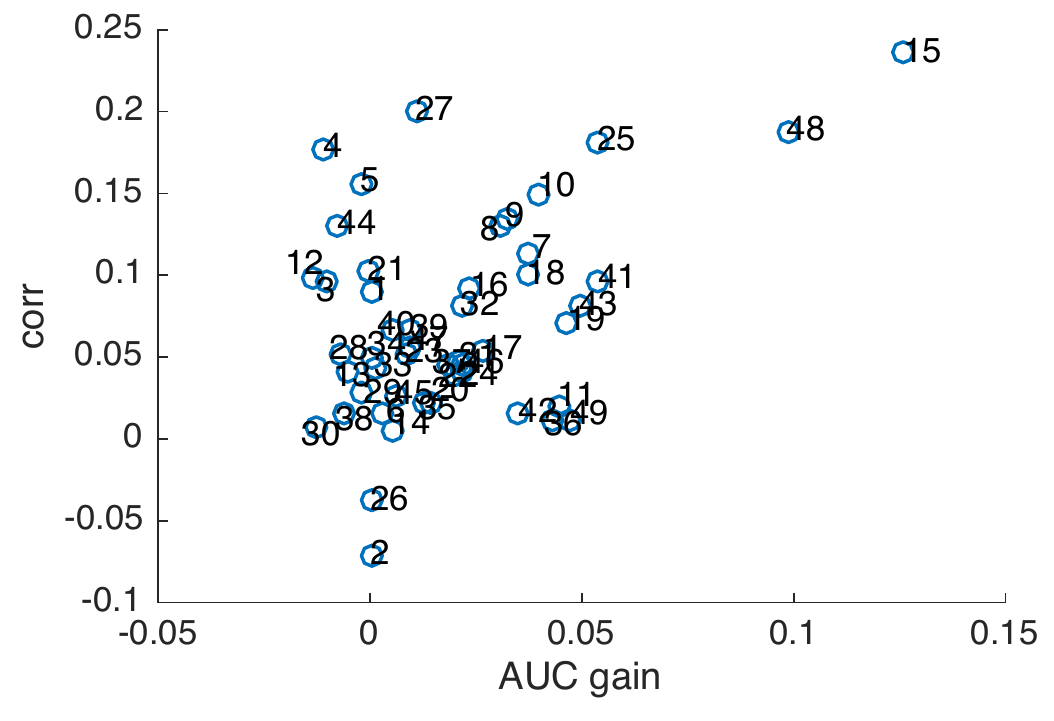}
\caption{\textbf{The higher the correlation between the data similarity and the decision table similarity, the higher the AUC gain.} }
\label{fig_gain_vs_corr}
\end{center}
\end{figure}

In this work, instances are clustered based on their similarities in the feature space. However, it is possible that this clustering may not be optimal in revealing model heterogeneity. A different view may be obtained by analyzing the instance-instance similarities in the model space. Therefore, we use the decision table, which describes the prediction performance of each model on each instance, to measure instance-instance similarity, and inspect whether the consistency between these two types of similarity measures can be predictive of the performance of ICE.  




Indeed, as shown in Figure \ref{fig_gain_vs_corr}, there exists a strong positive correlation (Pearson correlation coefficient = 0.425) between AUC gain and feature-model consistency, where the consistency is defined as the Pearson correlation correlation between the instance-instance similarities measured in the feature space and the instance-instance similarities measured using the decision table entries.
This result indicates the potential of improving our current work by feature selection and better clustering method on data $X$. Our intuition is that some features are more related with a classification task than the other features, and, we should be able to use these features for clustering for the classification task rather than use all the features. This also explains that the AUC gain on dataset 15-breast-cancer-1 (3 features, AUC gain = 0.126) is much larger than the AUC gain on dataset 16-breast-cancer-2 (50 features, AUC gain = 0.024). The three features of dataset 15-breast-cancer-1 are `tumor-size', `left or right breast' and `if irradiate', and, all the non-binary nominal features in the original breast cancer dataset from ~\cite{fernandez2014we} has been removed, while the dataset 16-breast-cancer-2 keeps all the other nominal features by One-Hot encoding. It is reasonable to imagine that the clustering on dataset 16 is influenced by some of the over-complicated and irrelevant features (for the classification task); therefore, the models built on those clusters are not optimized for the classification task. 
A potential future improvement is to cluster instances based on the output values from different models instead of on the feature values, or using both in an iterative manner. 

\section{Conclusion}
Based on the intuition that classifiers generated from different subdomains of training instances are needed in classification task, we proposed ICE, a novel multiple classifier generation and combination framework, which generally increases the diversity among submodels, and successfully associates the submodels to subdomains of instances. Evaluation results on 49 benchmarks show that our model has a stable improvement on a significant proportion of datasets over multiple existing MCS methods. A detailed component analysis shows that the different components of our algorithm work coordinately to achieve its performance. 
We believe that ICE can provide a novel choice of utilizing subdomain models to improve classification.

\section*{Acknowledgment}
This research was supported in part by grants from the National Science Foundation (award number IIS-1218201 and ABI-1565076), and the National Institutes of Health (award number G12MD007591 and U54CA217297).

\balance
\bibliography{My_Curr_Refs}

\bibliographystyle{IEEEtran}

\end{document}